# Word Embeddings for the Armenian Language: Intrinsic and Extrinsic Evaluation

Karen Avetisyan[1], Tsolak Ghukasyan[2]
Ivannikov Laboratory for System Programming at Russian-Armenian University, Yerevan, Armenia
[1]karavet@ispras.ru, [2]tsggukasyan@ispras.ru

## Abstract

In this work, we intrinsically and extrinsically evaluate and compare existing word embedding models for the Armenian language. Alongside, new embeddings are presented, trained using GloVe, fastText, CBOW, SkipGram algorithms. We adapt and use the word analogy task in intrinsic evaluation of embeddings. For extrinsic evaluation, two tasks are employed: morphological tagging and text classification. Tagging is performed on a deep neural network, using ArmTDP v2.3 dataset. For text classification, we propose a corpus of news articles categorized into 7 classes. The datasets are made public to serve as benchmarks for future models.

## Introduction

Dense representations of words have become an essential part of natural language processing systems. For many languages, there are various models available publicly for embedding a word to a continuous vector space, and choosing which model to use for a given task might often be problematic as embedding's performance strongly depends on the nature of the task. This work evaluates and compares the performance of word embedding models available for the Armenian language, aiming to provide insight into their performance on a diverse set of tasks. In addition, benchmark datasets and methods are established for evaluation of future models.

To the best of our knowledge, the first attempt to training dense representations for Armenian words was the 2015 SkipGram model by YerevaNN, trained on 2015 October 2 dump of Wikipedia. In 2017, Facebook released fastText models trained on Wikipedia for 90 languages [1], including Armenian. A year later, Facebook released another batch of fastText embeddings, trained on Common Crawl and Wikipedia [2]. Other publicly available embeddings include 4 models, released in 2018 as part of the pioNER project [3], using GloVe method [4] and trained on encyclopedia, fiction and news data. In addition to these 7 models, we present newly trained GloVe, fastText, CBOW embeddings. All mentioned models were included in our evaluation experiments.

Unfortunately, there is no single universal way of comparing the quality of word embeddings [5].

Available evaluation methods are grouped into 2 categories: (1) intrinsic, which test word relationships directly through word vectors, and (2) extrinsic, which test word vectors in downstream tasks such as morhological tagging, syntactic parsing, entity recognition, text categorization etc. In this work, for intrinsic evaluation we proceeded with word analogy questions. The Semantic-Syntactic Word Relationship test set (hereinafter referred to as "word analogy task") was first introduced in 2013 for English word vectors by Mikolov et al [6], then adapted for other languages for Czech [7], German [8], Italian [9], French, Hindi, Polish [2] etc. The are also analogy questions created for Armenian by YerevaNN, but these are not a direct adaptation of the word analogy task and test mostly semantic relationships[1]. In this work, we present and use a fully translated and adapted version of the original dataset.

In extrinsic evaluation, the performance of models varies depending on the nature of task (e.g. syntactic vs semantic). With that in mind, we selected 2 tasks for experiments: morphological tagging and classification of texts. The first is aimed to evaluate the applicability of models in morphology- and syntax-related tasks, while the second is mostly focused on measuring their ability to capture semantic properties. Morphological tagging experiments are performed on a deep neural network, using the ArmTDP dataset [10]. For text classification, we created a dataset of over 12000 news articles categorized into 7 classes: sport, politics, art, economy, accidents, weather, society.

The main contributions of this paper are (I) the word analogy task adapted for the Armenian language, (II) a corpus of categorized news texts, (III) GloVe, fastText, CBOW, SkipGram word representations, and (IV) performance evaluation of existing and proposed Armenian word embeddings. The datasets and embeddings are available on GitHub[2].

The first section of the paper describes existing and proposed embeddings. The second and third sections focus on the intrinsic and extrinsic evaluation respectively, providing the results of experiments at the end of each section.

# 1. Models

**1.1. Existing Models.** There already exist several trained models for Armenian.

*fastText Wiki*[3]: Published by Facebook in 2017, these embeddings were trained on Wikipedia, using SkipGram architecture [11] with window size 5, dimension 300, and character n-grams up to length 5.

*fastText CC*[4]: Published by Facebook in 2018, these embeddings were trained on Wikipedia and Common Crawl, using CBOW architecture with window size 5, dimension 300, and character n-grams up to length 5.

*SkipGram YerevaNN*[5]: Published by YerevaNN in 2015, these embeddings were trained on

---

[1] https://github.com/YerevaNN/word2vec-armenian-wiki/tree/master/analogies
[2] https://github.com/ispras-texterra/word-embeddings-eval-hy
[3] https://fasttext.cc/docs/en/pretrained-vectors.html
[4] https://fasttext.cc/docs/en/crawl-vectors.html
[5] https://github.com/YerevaNN/word2vec-armenian-wiki

Armenian Wikipedia using Skip-Gram with dimension 100.

*GloVe pioNER[6]*: Four models released by ISP RAS team, trained for 50, 100, 200 and 300 dimensions, using Stanford's GloVe algorithm with window size 15. These were trained on Wikipedia, parts of The Armenian Soviet Encyclopedia and EANC, news and blog texts.

### 1.2. Proposed Models

Apart from existing embeddings, we present 3 new models:

- 200-dimensional *GloVe* vectors, trained with window size 80;
- 300-dimensional *CBOW* and *SkipGram* vectors, trained with window size 5 and minimum word frequency threshold of 5;
- 200-dimensional *fastText* vectors, trained using SkipGram architecture, window size 3 and character n-grams up to length 3.

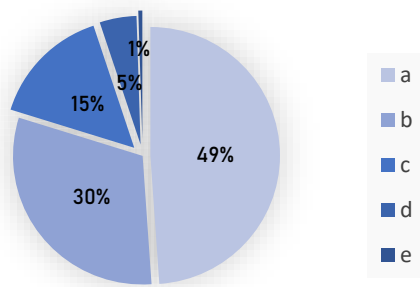

Figure 1. Distribution of Texts in Training Data.

The training data for these models was collected from various sources (Figure 1):

a. Wikipedia;
b. texts from news websites on the following topics: economics, international events, art, art, sports, law, politics, as well as blogs and interviews;
c. HC Corpora collected by Hans Christensen in 2011 from publicly available blogs and news, using a web crawler;
d. fiction texts taken from the open part of the EANC corpus [12];
e. digitized and reviewed part of Armenian soviet encyclopedia (as of February 2018) taken from Wikisource[7].

The texts were preprocessed by lowercasing all tokens and removing punctuation, digits. The final dataset consisted of 90.5 million tokens. Using the dataset, we trained word embeddings with GloVe, fastText, CBOW and Skip-gram methods. The specified hyperparameters were chosen based on the results on analogy questions. All fastText models come in .text and .bin formats. To try to take advantage of fastText's ability to generate vectors for out-of-vocabulary words, we separately tested .bin models as well.

## 2. Intrinsic Evaluation

To evaluate word vectors intrinsically we used an adaptation of the word analogy task introduced by Mikolov et al in 2013 for English word vectors. The original dataset contains 19544 semantic and syntactic questions divided into 14 sections. A question represents two pairs of words with the same analogy relationship: the first word is in the same relation with the second one as the third one with the fourth. Sections are divided into semantic and syntactic categories: first 5 sections test semantic analogies, and the rest syntactic.

In order to make a similar set for our experiments,

---

[6] https://github.com/ispras-texterra/pioner

[7] https://hy.wikisource.org/wiki/ Հայկական_սովետական_հանրագիտարան

we mostly directly translated original's examples into Armenian, using dictionaries[8,9] and Wikipedia[10] as a reference.

Several issues arose during translation. In some country-capital pairs, such as *Algeria - Algiers,* the country and its capital have the same translation (*Ալժիր*), thus all lines including such pairs were removed in order to restrict the task strictly to solving analogies. In other sections, the examples that didn't have distinct Armenian translation (e.g. *he-she, his-her*), were replaced by a different pair of words. Some country or city names like *Thailand* have multiple translations in Armenian (*Թաիլանդ, Թայլանդ*), and Wikipedia was used to resolve this kind of situations. To make a localized adaptation, the *city-in-state* category was replaced by regions of Armenia and their capitals, similar to adaptations for other languages [2]. The *comparative* section was fully removed because comparative adjectives in Armenian are multi-word expressions. In the *present-participle* section, the participles in *-ող* (ենթակայական) were used. While translating country-nationality analogies, we used corresponding demonyms instead of the ethnic majority's name.

Table 1. Sections of Word Analogy Task.

| Section | Questions | Type |
|---|---|---|
| *capital-common-countries* | 506 | Semantic |
| *capital-world* | 4369 | Semantic |
| *currency* | 866 | Semantic |
| *city-in-state* | 56 | Semantic |
| *family* | 506 | Semantic |
| *gram1-adjective-to-adverb* | 992 | Syntactic |
| *gram2-opposite* | 812 | Syntactic |
| *gram3-superlative* | 1122 | Syntactic |
| *gram4-present-participle* | 1056 | Syntactic |
| *gram5-nationality-adjective* | 1599 | Syntactic |
| *gram6-past-tense* | 1560 | Syntactic |
| *gram7-plural* | 1332 | Syntactic |
| *gram8-plural-verbs* | 870 | Syntactic |

Table 2. Section-Wise Accuracy of Embeddings on Word Analogy Task.

| Models | capital common | capital world | curr. | city-state | family | gram1 | gram2 | gram3 | gram4 | gram5 | gram6 | gram7 | gram8 |
|---|---|---|---|---|---|---|---|---|---|---|---|---|---|
| fastText [Wiki .text] | 5.34% | 0.78% | 0% | 0% | 4.54% | 14.91% | 30.29% | 39.57% | 7.19% | 44.21% | 23.71% | 29.35% | 0.45% |
| fastText [Wiki .bin] | 5.34% | 0.77% | 0% | 0% | 4.54% | 16.53% | 30.29% | 27.98% | 7.19% | 47.52% | 23.71% | 29.35% | 0.45% |
| fastText [CC .text] | 32.61% | 11.42% | 2.77% | 7.14% | 13.83% | 22.07% | 30.66% | 43.76% | 4.45% | 41.58% | 18.33% | 19.96% | 5.51% |
| fastText [CC .bin] | 72.53% | 39.28% | **11.55%** | **48.21%** | **47.83%** | 25.2% | 36.21% | 49.64% | 19.7% | 8.53% | 23.08% | 41.89% | **41.03%** |
| fastText [new .text] | 27.66% | 8.1% | 0.1% | 1.79% | 16.2% | 28.02% | **41.74%** | 48.3% | **23.95%** | 54.59% | 50.51% | 53.67% | 6.09% |
| fastText [new .bin] | 27.67% | 8.1% | 1.03% | 1.79% | 16.2% | **30.14%** | **41.74%** | 58.82% | **23.95%** | 54.59% | 50.51% | 53.67% | 6.09% |
| SkipGram [YerevaNN] | 39.32% | 10.66% | 2.07% | 8.93% | 5.73% | 4.03% | 0.61% | 3.74% | 1.23% | 23.57% | 0.12% | 5.78% | 0.8% |
| SkipGram [new] | 36.17% | 17.37% | 2.3% | 3.57% | 17.79% | 7.56% | 12.43% | 16.39% | 1.7% | 37.77% | 4.93% | 16.81% | 10.34% |
| CBOW [new] | 28.65% | 13.04% | 1.5% | 5.36% | 29.05% | 10.48% | 14.77% | 17.91% | 5.49% | 24.26% | 6.98% | 28.6% | 11.83% |
| GloVe [pioNER dim50] | 8.1% | 1.06% | 0.11% | 0% | 6.71% | 3.32% | 2.46% | 3.29% | 0.94% | 11.75% | 1.02% | 6.98% | 1.26% |
| GloVe [pioNER dim100] | 10.67% | 1.67% | 0.46% | 3.57% | 10.27% | 4.53% | 5.41% | 4.72% | 1.51% | 16.51% | 1.15% | 7.43% | 3.9% |
| GloVe [pioNER dim200] | 10.67% | 2.28% | 0.8% | 7.14% | 11.66% | 3.52% | 8.74% | 7.13% | 1.32% | 15.69% | 1.02% | 5.7% | 2.87% |
| GloVe [pioNER dim300] | 10.87% | 2.05% | 0.46% | 5.36% | 11.46% | 3.22% | 7.88% | 5.79% | 0.94% | 13.75% | 0.7% | 4.72% | 1.49% |
| GloVe [new] | **75.3%** | **49.14%** | 2.19% | 23.21% | 15.8% | 6.55% | 11.2% | 12.74% | 2.27% | 47.71% | 1.85% | 20.49% | 5.4% |

---

[8] https://bararanonline.com
[9] https://hy.wiktionary.org
[10] https://hy.wikipedia.org

Table 3. Total and Section-Wise Average Accuracy of Embeddings on Word Analogy Task.

| Model | Semantic | Syntactic | Total | Semantic (avg.) | Syntactic(avg.) | Total (avg.) |
|---|---|---|---|---|---|---|
| fastText [Wiki .text] | 1.33% | 24.88% | 15.39% | 2.13% | 22.87% | 14.89% |
| fastText [Wiki .bin] | 1.33% | 25.53% | 15.78% | 2.13% | 23.71% | 15.41% |
| fastText [CC .text] | 12.08% | 24.3% | 19.38% | 13.55% | 23.29% | 19.54% |
| fastText [CC .bin] | 38.9% | 35.96% | **37.14%** | 43.88% | 35.65% | **38.81%** |
| fastText [new .text] | 9.29% | 41.11% | 28.29% | 10.77% | 38.35% | 27.74% |
| fastText [new .bin] | 9.3% | **42.48%** | 29.11% | 10.95% | **39.93%** | 28.79% |
| SkipGram [YerevaNN] | 11.37% | 5.98% | 8.15% | 13.34% | 4.98% | 8.19% |
| SkipGram [new] | 16.72% | 14.69% | 15.51% | 15.44% | 13.49% | 14.24% |
| CBOW [new] | 13.92% | 15.66% | 14.96% | 15.52% | 15.04% | 15.22% |
| GloVe [pioNER dim50] | 1.93% | 4.36% | 3.38% | 3.19% | 3.87% | 3.61% |
| GloVe [pioNER dim100] | 2.93% | 6.13% | 4.84% | 5.33% | 5.64% | 5.52% |
| GloVe [pioNER dim200] | 3.55% | 6.07% | 5.06% | 6.51% | 5.74% | 6.04% |
| GloVe [pioNER dim300] | 3.33% | 5.11% | 4.39% | 6.04% | 4.81% | 5.28% |
| GloVe [new] | **41.88%** | 15.35% | 26.04% | 33.12% | 13.52% | 21.06% |

The obtained test set has 15646 questions and 13 sections (Table 1). Similar to the original, translated word analogy task is not balanced in terms of the number of questions per category and the sizes of semantic and syntactic parts. For example, country-capital relations (e.g. Աթենք ~ Հունաստան = Կահիրե ~ Եգիպտոս) comprise over 70% of semantic relations.

The accuracy of vectors on a question is determined as follows ($WV_i$ denotes the word vector of $i^{th}$ word in the question):

I. Calculate $p = WV_2 - WV_1 + WV_3$
II. Check $p \approx WV_4$ (this holds if p is closest to 4$^{th}$ word's vector among all words)

If the equation in II holds, the model is said to predict the question correctly.

**Results**. To evaluate the accuracy of vectors, we used gensim's *evaluate_word_analogies* function[11]. While computing the accuracy, the vocabulary of models was restricted to 400000 words. Accuracy is calculated on each section (Table 2) and all examples, alongside we provide average accuracy over all sections (Table 3). Proposed fastText and GloVe embeddings correctly predicted highest number of syntactic and semantic examples respectively. However, GloVe's score is clearly inflated by its performance on country-capital examples. On average, Facebook's Common Crawl vectors produced higher results in semantic sections and in total over all sections. As evidenced by the marked difference in accuracy between .bin and .text formats, the ability of .bin to generate vectors for out-of-vocabulary words greatly boosted its performance.

---

[11] github.com/RaRe-Technologies/gensim/blob/develop/gensim/models/keyedvectors.py

# 3. Extrinsic Evaluation

The second way of comparing the quality of vectors is to use them in downstream tasks. The comparison's outcome depends on the nature of these tasks and settings [5]. For that purpose, we performed experiments on 2 different tasks: morphological tagging and text classification.

**3.1. Morphological Analysis.** To check the quality of vectors in a morphological task, we trained and evaluated a neural network-based tagger on ArmTDP treebank, using word embeddings as input. The treebank's sentences are morphologically annotated, and the tagger's task was to predict the following 2 fields:

1. UPOS: the universal part-of-speech tag.
2. FEATS: the list of morphological features (number, case, animacy etc).

For the tagger, we used a neural network with 1 sentence-level bidirectional recurrent LSTM unit (Fig. 2). The pre-trained embeddings were used as input to the network. Apart from word embeddings, the network's input included character-based features extracted through a convolutional layer. The tagger was jointly trained to predict 2 tags for each input token: its UPOS and the concatenation of FEATS tags.

Figure 2. Morphological Tagger's Neural Network.

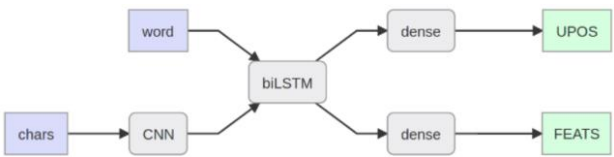

As of v2.3, ArmTDP provides only train and test sets. We randomly detached 20% of the original train set into a development set and used the other 80% to train the network for 200 epochs, saving parameters after each epoch. For optimization we used gradient descent with initial learning rate of 0.6 and time-based decay of 0.05. For testing, the model with highest accuracy on the development set was used. The backpropagation of errors to the word embeddings was blocked during training.

Table 4. Taggers' Accuracy on ArmTDP v2.3 Test Set.

| Model | Dev* | | Test | |
|---|---|---|---|---|
| | UPOS | FEATS | UPOS | FEATS |
| fastText [Wiki .text] | 92.99% | 83.83% | 89.54% | 81.03% |
| fastText [Wiki .bin] | 89.27% | 75.96% | 84.35% | 71.14% |
| fastText [CC .text] | 91.86% | 79.85% | 88.38% | 75.44% |
| fastText [CC .bin] | 91.54% | 77.78% | 87.59% | 73.29% |
| fastText [new .text] | **94.64%** | 85.89% | **93.35%** | 83.99% |
| fastText [new .bin] | 91.43% | 80.1% | 87.55% | 74.78% |
| SkipGram [YerevaNN] | 91.44% | 80.04% | 87.45% | 75.68% |
| SkipGram [new] | 93.9% | **86.45%** | 91.6% | 83.77% |
| CBOW [new] | 93.87% | 85.39% | 92.72% | **84.07%** |
| GloVe [pioNER dim50] | 91.78% | 82.42% | 89.12% | 80.18% |
| GloVe [pioNER dim100] | 91.78% | 82.95% | 89.06% | 80.51% |
| GloVe [pioNER dim200] | 92.34% | 83.77% | 88.90% | 80.07% |
| GloVe [pioNER dim300] | 91.70% | 83.19% | 89.09% | 80.78% |
| Glove [new] | 94.09% | 86.23% | 92.98% | 83.97% |

*Dev refers to the 20% of original train set that we used for validation

**Results.** The experiments for all embeddings were carried out for 10 random seeds and then averaged. The results are reported in Table 4. Facebook's fastText models performed poorly in comparison to others. A striking observation is that in contrast to analogies, here fastText .text vectors outperformed .bin by a big margin, especially on FEATS. Overall, proposed models demonstrated highest accuracy. The best was fastText, followed by CBOW and GloVe.

**3.2. Text Classification.** In order to check the quality of word vectors in a classification task, we collected 12428 articles from ilur.am news website[12] and categorized into 7 topics: art, economy, sport, accidents, politics, society and weather (2667299 tokens in total).

The dataset's texts' 80% were used as train data and the other 20% as test data. All texts were preprocessed by removing stop words[13]. As features for the classifier the average of a text's word vectors was used with tf-idf weighting. Then, one-vs-rest logistic regression classifier with liblinear solver was applied.

Figure 3. Categories in News Texts Dataset.

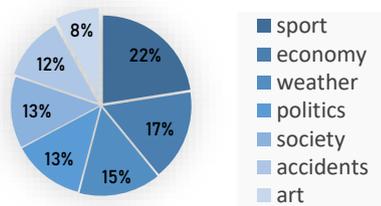

Table 5. Accuracy, Precision, Recall, F-1 (macro) Classification Scores on News Texts Dataset.

| Model | Accuracy | Precision | Recall | F1 |
|---|---|---|---|---|
| fastText [Wiki .text] | 66.63 | 60.38 | 63.02 | 58.96 |
| fastText [Wiki .bin] | 65.79 | 59.78 | 63.97 | 58.15 |
| fastText [CC .text] | 65.75 | 59.15 | 63.68 | 57.03 |
| fastText [CC .bin] | 65.51 | 58.96 | 63.43 | 56.94 |
| fastText [new .text] | 63.34 | 56.97 | 59.81 | 55.18 |
| fastText [new .bin] | 60.12 | 53.66 | 55.4 | 51.96 |
| SkipGram [YerevaNN] | 64.34 | 57.87 | 59.73 | 56.28 |
| SkipGram [new] | 66.68 | 60.84 | 63.56 | 59.83 |
| CBOW [new] | 67.92 | 61.94 | 65.2 | 60.94 |
| GloVe [pioNER dim50] | 64.26 | 57.57 | 58.39 | 54.4 |
| GloVe [pioNER dim100] | 65.91 | 60.15 | 62.34 | 58.91 |
| GloVe [pioNER dim200] | 68.16 | 62.13 | 65.54 | 60.6 |
| GloVe [pioNER dim300] | 67.85 | 61.7 | 65.36 | 60.43 |
| GloVe [new] | **69.77** | **63.93** | **66.55** | **63.13** |

**Results.** The accuracy, precision, recall and F1 scores shown in Table 5, were computed for each model. Overall, proposed GloVe vectors achieved highest scores in all metrics, outperforming fastText, SkipGram, CBOW. Somewhat surprisingly, high-dimensional GloVe vectors from pioNER project, which performed poorly on analogies, are among the top performers here. It is also worthy to note that among fastText models, .bin vectors again produced worse scores than .text.

## Discussion and Conclusion

This work evaluates and compares the performance of publicly available word embeddings for the Armenian language in 3 different tasks. To that end, benchmark datasets are presented. For intrinsic tests, the word analogy task is translated and adapted. For extrinsic evaluation, a neural network-based morphological tagger is employed, and a corpus of news texts is created for comparison of performance in a classification task. Alongside, new embeddings are released that outperform most of the existing models on these tasks.

Generally, fastText models produced better results in morphosyntactic tasks, while GloVe models performed better in tasks sensitive to semantic information. On analogy questions, Facebook's Common Crawl vectors were the winner by a big margin, but performed relatively poorly in morphological tagging and text classification tasks. In latter tasks, proposed CBOW, fastText, and GloVe models demonstrated noticeably higher accuracy. Overall, the results also illustrate that accuracy on analogy questions is not a good indicator of embeddings' performance on downstream tasks.

---

[12] http://www.ilur.am/

[13] https://github.com/stopwords-iso/stopwords-hy